\documentclass[letterpaper]{article} 
\usepackage{aaai2027}  
\usepackage[hyphens]{url}  
\usepackage{graphicx} 
\graphicspath{{figures/}{Figures/}}
\usepackage{natbib}  
\setcitestyle{numbers}
\usepackage{caption} 
\usepackage{booktabs}
\usepackage{amsmath}
\usepackage{amsfonts}
\usepackage{multirow}
\usepackage{enumitem}
\title{SKIMIX: Multi-Agent Harness-Time Scaling with \\ Skill Mixture for Dynamic Harness Engineering}
\author{Jia Luo}
\affiliations{
    Huazhong University of Science and Technology\\
    u202317016@hust.edu.cn
}

\begin{document}

\maketitle

\begin{abstract}
AI agents increasingly rely on large skill libraries, but selecting, combining, and maintaining skills remains difficult. We propose SKIMIX, a multi-agent framework in which agents with different skill portfolios collaborate through iterative refinement. SKIMIX combines embedding-based skill retrieval, submodular anti-dilution routing, and adaptive skill evolution. Across six reasoning benchmarks, multi-agent collaboration substantially improves open-ended mathematical reasoning but offers limited or negative gains on multiple-choice tasks. Agent-count scaling is non-monotonic, and most improvements arise during the first refinement round. These results show that task characteristics determine whether skill-level ensembles help and provide practical guidance for scalable agent design.

\end{abstract}


\section{Introduction}
\label{sec:intro}

The rapid advancement of Large Language Models (LLMs) has given rise to AI agent harnesses: software frameworks that equip a base LLM with a curated library of \textit{skills}. A skill can be a specialized prompt, a tool integration, a reasoning strategy, or a domain-specific workflow~\cite{wu2024autogen,li2023camelcommunicativeagentsmind}. Modern agent platforms (Claude Code, ChatGPT Agents, Gemini, AutoGen) now ship with tens to hundreds of pre-built skills covering code interpretation, web search, multi-modal analysis, document understanding, and domain-specific reasoning.

\begin{figure*}[t]
    \centering
    \includegraphics[width=1\textwidth,trim=2cm 0 2cm 0,clip]{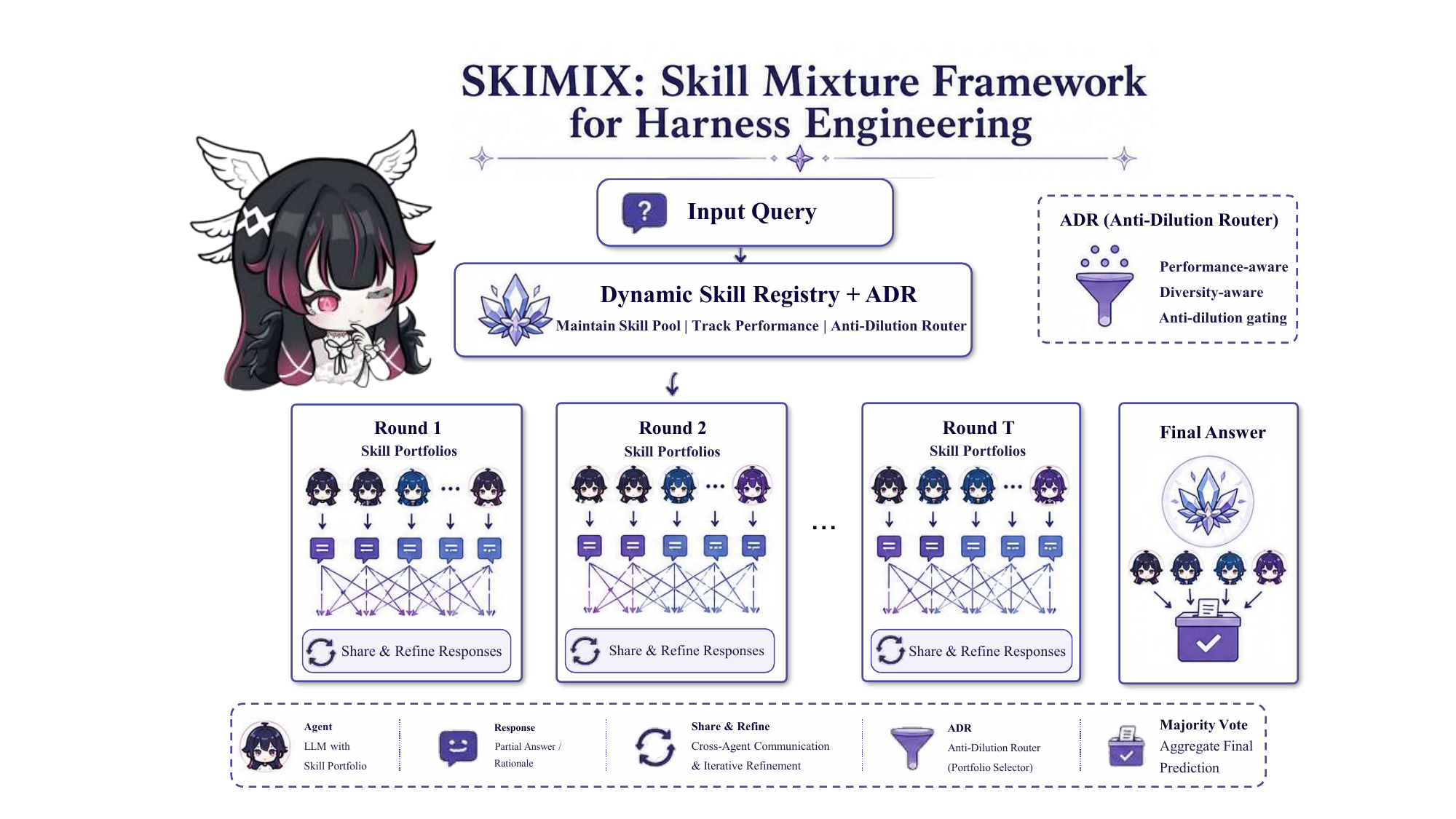}
    \caption{Overview of the SKIMIX framework. Agents with diverse skill portfolios process queries in parallel; responses are shared and refined across rounds. The Dynamic Skill Registry and Anti-Dilution Router manage portfolio selection.}
    \label{fig:skimix_overview}
\end{figure*}

Two problems stand out. \textbf{First}, for any given query, nobody knows the best combination of skills. Queries rarely telegraph which skills they need, and the space of possible skill portfolios is too large to search exhaustively. Current systems either stick to a single fixed pipeline or ask the user to pick skills manually. Both approaches leave most of the skill library unused. \textbf{Second}, as skill libraries grow, \textit{skill dilution} becomes a real issue. Redundant skills add noise, make routing harder, and pull down ensemble performance~\cite{shazeer2017outrageouslylargeneuralnetworks}. No existing system handles both scalable composition and anti-dilution together.

We propose \textbf{Skill Mixture (SKIMIX)} to tackle both. The idea is straightforward: run multiple agents in parallel, give each one a different \textit{skill portfolio} (a curated subset of the skill library with specific invocation strategies), and let them iteratively share and refine answers across rounds. The design borrows from test-time scaling methods like Mixture-of-Agents (MoA)~\cite{wang2024mixtureofagentsenhanceslargelanguage} and Tool-Use Mixture (TUMIX)~\cite{chen2025tumixmultiagenttesttimescaling}, but where those papers focus on model diversity or tool diversity, we focus on \textit{skill composition}. That distinction matters because skills are cheaper to add, easier to mix, and more directly tied to task performance than models or tools.

SKIMIX adds three mechanisms that go beyond the basic mixture framework:

\begin{enumerate}[leftmargin=*,itemsep=0.5pt,topsep=0pt]
    \item \textbf{Dynamic Skill Registry (DSR)}: A learned embedding space where each skill gets a dense vector encoding its capability profile, invocation patterns, and success statistics. Skills are retrieved by similarity-based routing rather than hard-coded rules.
    \item \textbf{Anti-Dilution Routing (ADR)}: As new skills enter the registry, ADR computes each one's \textit{marginal diversity contribution} via submodular maximization. Skills that do not add enough diversity are down-weighted or archived, which stops dilution before it starts.
    \item \textbf{Adaptive Skill Evolution (ASE)}: Skills are continuously scored on recent performance. Stale ones get deprecated. When usage patterns reveal gaps, the system synthesizes new skills from observed reasoning traces.
\end{enumerate}

We evaluate SKIMIX on six reasoning benchmarks (AIME, GPQA Diamond, HLE, MMLU-Pro, MATH-500, and BBH). The main findings are as follows:

\begin{itemize}[leftmargin=*,itemsep=0.5pt,topsep=0pt]
    \item Task type determines whether more agents help or hurt. On open-ended math, multi-agent diversity adds +33\% over self-refinement. On MCQ tasks, the best method is single-agent refinement (88.0\% on GPQA), and adding agents consistently degrades accuracy. Mixed-domain benchmarks sit in between, with small gains either way.
    \item On AIME, base accuracy starts at \textbf{0.0\%}, jumps to \textbf{40.0\%} with self-refinement, and reaches \textbf{73.3\%} (3 agents) and \textbf{76.7\%} (15 agents) with SKIMIX.
    \item Agent scaling is not monotonic: 3 agents beat 5 on AIME (73.3\% vs. 70.0\%) and GPQA (78.0\% vs. 74.0\%). On MCQ benchmarks, every additional agent makes things worse.
    \item Round 2 is where most of the accuracy gain happens, across all methods and benchmarks. Round 3 adds little and sometimes subtracts, which directly motivates adaptive early stopping.
\end{itemize}

We believe this is the first systematic study of skill-level ensemble scaling across diverse task types, the first to document the MCQ/open-ended divergence in multi-agent settings, and the first to address skill dilution through principled submodular optimization.

\section{Related Work}
\label{sec:related}

\subsection{AI Agent Harnesses and Skill Frameworks}

Modern AI agent harnesses organize capabilities into modular ``skills''---self-contained units combining prompts, tool specifications, and execution logic. Frameworks such as AutoGen~\cite{wu2024autogen}, LangChain, CrewAI, and Claude Code's skill system exemplify this paradigm. Research has explored agent specialization~\cite{hong2024metagptmetaprogrammingmultiagent,chen2023agentversefacilitatingmultiagentcollaboration}, multi-agent collaboration~\cite{li2023camelcommunicativeagentsmind,zhang2024more,wang2024mixtureofagentsenhanceslargelanguage}, and workflow optimization~\cite{qian2025toolrlrewardtoollearning}. However, existing work treats skills as static resources; no prior work addresses \textit{skill composition at scale}---dynamically selecting, combining, and managing skills from a growing library while preventing dilution.

\subsection{Test-Time Scaling and Multi-Agent Reasoning}

Test-time scaling improves LLM performance by allocating additional compute at inference. Approaches include repeated sampling with majority voting~\cite{brown2024largelanguagemonkeysscaling}, self-reflection~\cite{madaan2023self,shinn2023reflexion}, and multi-agent frameworks~\cite{shazeer2017outrageouslylargeneuralnetworks,zhang2024more}. TUMIX~\cite{chen2025tumixmultiagenttesttimescaling} introduces tool-use diversity within a single LLM, demonstrating that agent diversity and quality are critical for test-time scaling. These methods focus on \textit{model} or \textit{tool} diversity. SKIMIX focuses on \textit{skill portfolio} diversity and introduces mechanisms for managing skill ecosystems at scale.

\subsection{Skill Dilution and Information Overload}

As skill libraries expand, selecting the most relevant skills becomes increasingly challenging due to information overload and redundancy. Similar phenomena have been observed in ensemble learning, where incorporating weak or redundant models can degrade overall performance~\cite{dietterich2000ensemble}, and in long-context language modeling, where irrelevant context distracts reasoning and reduces model effectiveness~\cite{liu2023lostmiddlelanguagemodels}. To mitigate redundancy while preserving diversity, submodular optimization has been widely studied for subset selection in machine learning and information retrieval~\cite{krause2014submodular}. Building on these insights, SKIMIX formulates skill routing as a submodular maximization problem, selecting a compact yet complementary subset of skills to alleviate skill dilution and improve reasoning efficiency.

\section{Skill Mixture (SKIMIX)}
\label{sec:method}

\subsection{Pre-Designed Diverse Skill Portfolios}
\label{sec:portfolios}

SKIMIX is formalized as sequential decision-making under a compute budget with diverse and correlated skill portfolios. Let $q$ be a task with unknown correct answer $a^\star$ in answer space $\mathcal{A}$. There is a global skill registry $\mathcal{S} = \{s_1, \ldots, s_K\}$ where each skill $s_i$ has a learned embedding $\mathbf{e}_i \in \mathbb{R}^d$. A \textit{skill portfolio} $P_j \subseteq \mathcal{S}$ is a subset of skills assigned to agent $j$, with competence $p_j(q) = P\{Y_j = a^\star \mid q\}$.

A policy $\pi$ chooses in each round: (i) which portfolios to activate, (ii) the communication graph, (iii) the stopping rule, and (iv) the aggregation rule. The objective is:

\begin{equation}
\max_{\pi} \; P\{\hat{a}^\pi = a^\star\} - \lambda \cdot \text{Cost}_\pi,
\label{eq:objective}
\end{equation}

where $\lambda > 0$ trades off compute and accuracy. In the default SKIMIX setting, we utilize 15 pre-designed skill portfolios spanning five categories: reasoning (Base, CoT, StructuredReason, Memory, Planning), code (CoTcode, Code, Code+), search (Search with three backends), hybrid (DualMode, Guided, Guided+), and specialized domains (Analysis, Verification, MultiModal).

\subsection{Dynamic Skill Registry and Anti-Dilution Routing}
\label{sec:registry}

\subsubsection{Dynamic Skill Registry (DSR)}

The DSR maintains embeddings $\mathbf{e}_i \in \mathbb{R}^d$ for each skill, initialized from skill descriptions via sentence encoding and updated via usage feedback:

\begin{equation}
\mathbf{e}_i^{(t+1)} = \mathbf{e}_i^{(t)} + \eta \cdot \nabla_{\mathbf{e}_i} \mathcal{L}_{\text{align}}(q, \mathbf{e}_i, y).
\label{eq:embedding_update}
\end{equation}

The registry supports registration (new skills embedded with k-NN indexing), retrieval (top-$k$ by cosine similarity), and evolution (decaying skills flagged for deprecation or regeneration).

\subsubsection{Anti-Dilution Routing (ADR)}

As the skill library grows, ADR selects portfolios via submodular maximization. Given query $q$, we seek $P \subseteq \mathcal{S}$ of size $k$ maximizing:

\begin{equation}
P^* = \arg\max_{P \subseteq \mathcal{S}, |P| \leq k} \underbrace{\sum_{s \in P} \text{rel}(s, q)}_{\text{Relevance}} + \gamma \cdot \underbrace{\log \det(\mathbf{I} + \mathbf{K}_P)}_{\text{Diversity}},
\label{eq:adr}
\end{equation}

where $\text{rel}(s, q)$ is embedding similarity and $\mathbf{K}_P$ is the skill similarity kernel. The log-determinant encourages portfolios with low inter-skill similarity, directly countering dilution. The greedy solution achieves a $(1-1/e)$ approximation. Each skill's marginal gain is $\Delta(s \mid P) = F(P \cup \{s\}; q) - F(P; q)$. Skills with $\Delta(s \mid P) < \tau_{\text{min}}$ are excluded or archived.

\subsection{Refinement as Message Passing}
\label{sec:refinement}

In each round, every agent independently generates a new solution by considering both the original query and all solutions from the previous round. We evaluate using two metrics: \textbf{average accuracy} and \textbf{coverage} (probability of at least one correct answer):

\begin{equation}
\text{Coverage}(\mathcal{P}) = P\left( \bigcup_{j=1}^{m} \{Y_j = a^\star\} \right).
\label{eq:coverage}
\end{equation}

Our real experiments (Sec.~\ref{sec:experiments}) reveal the refinement dynamics: coverage and accuracy both increase sharply from Round 1 to Round 2, then plateau or slightly decline, indicating that correct answers can be discarded with excessive refinement.

\subsection{Termination and Answer Selection}
\label{sec:termination}

Let $A_r$ denote accuracy after round $r$. The expected marginal value is $\Delta_r = \mathbb{E}[A_{r+1} - A_r \mid \text{signals}]$. We stop when $\Delta_r \leq \lambda \cdot \text{marginal cost}$. Our \textbf{LLM-as-Judge} strategy queries the LLM to decide termination, with a minimum of 2 rounds. After termination, the final answer is selected via majority voting.

\section{Experiments}
\label{sec:experiments}

\subsection{Experimental Settings}
\label{sec:settings}

\textbf{Benchmarks.}
We evaluate six benchmarks in three settings: open-ended reasoning (AIME 2024\&2025 and MATH-500), multiple-choice reasoning (GPQA Diamond~\cite{rein2023gpqagraduatelevelgoogleproofqa}, MMLU-Pro~\cite{wang2024mmlu}, and BBH~\cite{srivastava2023beyond}), and mixed-format reasoning (HLE).

\textbf{Methods.}
We compare five inference configurations:
\textbf{Base} (single-pass),
\textbf{Self-Refine} (1 agent, 3 rounds),
and \textbf{SKIMIX} with 3, 5, or 15 agents.
Each SKIMIX agent is assigned a distinct skill portfolio from a predefined library, and agent outputs are shared and aggregated after each round. Due to computational cost, SKIMIX-15 is evaluated only on AIME, GPQA, and HLE.

\textbf{Implementation.}
All methods use \texttt{DeepSeek-V3.2} with identical decoding settings (temperature 0.7 and maximum 2048 output tokens).

\textbf{Evaluation.}
Performance is measured by exact answer matching, following the official evaluation protocol for each benchmark. We additionally report per-round \textbf{agent accuracy} and \textbf{coverage}, defined as the probability that at least one agent produces the correct answer (Eq.~\ref{eq:coverage}).
\subsection{Overall Results}
\label{sec:overall}

\begin{table*}[t]
    \centering
    \resizebox{\textwidth}{!}{%
    \begin{tabular}{lcccccc}
        \toprule
        \textbf{Method} & \textbf{AIME} & \textbf{GPQA} & \textbf{HLE} & \textbf{MMLU-Pro} & \textbf{MATH-500} & \textbf{BBH} \\
        \midrule
        Base          & 0.0\%  & 64.0\% & 8.0\%  & 6.0\%  & 16.0\% & 10.4\% \\
        Self-Refine   & 40.0\% & \textbf{88.0\%} & 43.0\% & 51.0\% & 61.0\% & 50.7\% \\
        SKIMIX-3      & 73.3\% & 78.0\% & 45.0\% & 49.0\% & 71.0\% & 59.7\% \\
        SKIMIX-5      & 70.0\% & 74.0\% & 47.0\% & \textbf{52.0\%} & \textbf{72.0\%} & \textbf{62.7\%} \\
        SKIMIX-15     & \textbf{76.7\%} & 72.0\% & \textbf{51.0\%} & --     & --     & -- \\
        \midrule
       \textit{Trend} & $\sim$   & $\downarrow$ & $\uparrow$ & $\sim$    & $\uparrow$ & $\uparrow$ \\
        \bottomrule
    \end{tabular}%
    }
    \caption{Accuracy (\%) across six benchmarks and five methods. Bold indicates best per benchmark. Arrows indicate trend with increasing agents: $\uparrow$ = monotonic improvement, $\downarrow$ = monotonic decline, $\sim$ = non-monotonic or flat. Dash (--) indicates not evaluated due to compute constraints. All results from real API calls to DeepSeek-V3.2.}
    \label{tab:main_results}
\end{table*}

Table~\ref{tab:main_results} presents the full results across all six benchmarks. Three qualitatively distinct patterns emerge, defining the central finding of this paper:

\smallskip
\noindent\textbf{Pattern 1---Multi-agent dominance (AIME, MATH):} On open-ended mathematical reasoning, multi-agent SKIMIX substantially outperforms single-agent self-refinement. On AIME, self-refinement reaches 40.0\% while SKIMIX-3 achieves 73.3\% (+33.3\% absolute, +83\% relative) and SKIMIX-15 reaches 76.7\%. On MATH-500, SKIMIX-5 achieves 72.0\% vs.\ self-refinement's 61.0\% (+11.0\%). The base model's near-zero AIME accuracy (0.0\%) underscores the benchmark's extreme difficulty, making the 76.7\% final result a 76.7-point absolute gain over the single-pass baseline.

\smallskip
\noindent\textbf{Pattern 2---Single-agent superiority (GPQA):} On GPQA Diamond, a rigorous MCQ science benchmark, single-agent self-refinement achieves the best overall accuracy of 88.0\%. Adding more agents with diverse skill portfolios \textit{degrades} performance: SKIMIX-3 = 78.0\% ($-$10.0\%), SKIMIX-5 = 74.0\% ($-$14.0\%), SKIMIX-15 = 72.0\% ($-$16.0\%). The monotonic decline with increasing agents indicates that on well-structured MCQ tasks, diverse reasoning paths introduce conflicting signals that undermine answer selection, whereas a single agent maintaining self-consistency across rounds achieves optimal calibration.

\smallskip
\noindent\textbf{Pattern 3---Flat regime (HLE, MMLU-Pro, BBH):} On HLE and MMLU-Pro, the gains from both refinement and multi-agent diversity are modest. HLE improves from 8.0\% (base) to 43.0\% (self-refine) and plateaus at 51.0\% (SKIMIX-15), reflecting the benchmark's intrinsic difficulty. MMLU-Pro shows minimal sensitivity to method: self-refine = 51.0\%, SKIMIX-3 = 49.0\%, SKIMIX-5 = 52.0\%---a range of only 3 points. BBH shows a moderate improvement trend (50.7\% $\rightarrow$ 62.7\%) but the gains are substantially smaller than on open-ended reasoning tasks.

\subsection{Round-by-Round Refinement Dynamics}
\label{sec:rounds}

\begin{figure*}[t]
    \centering
    \begin{minipage}[t]{0.32\textwidth}
        \centering
        \includegraphics[width=\textwidth]{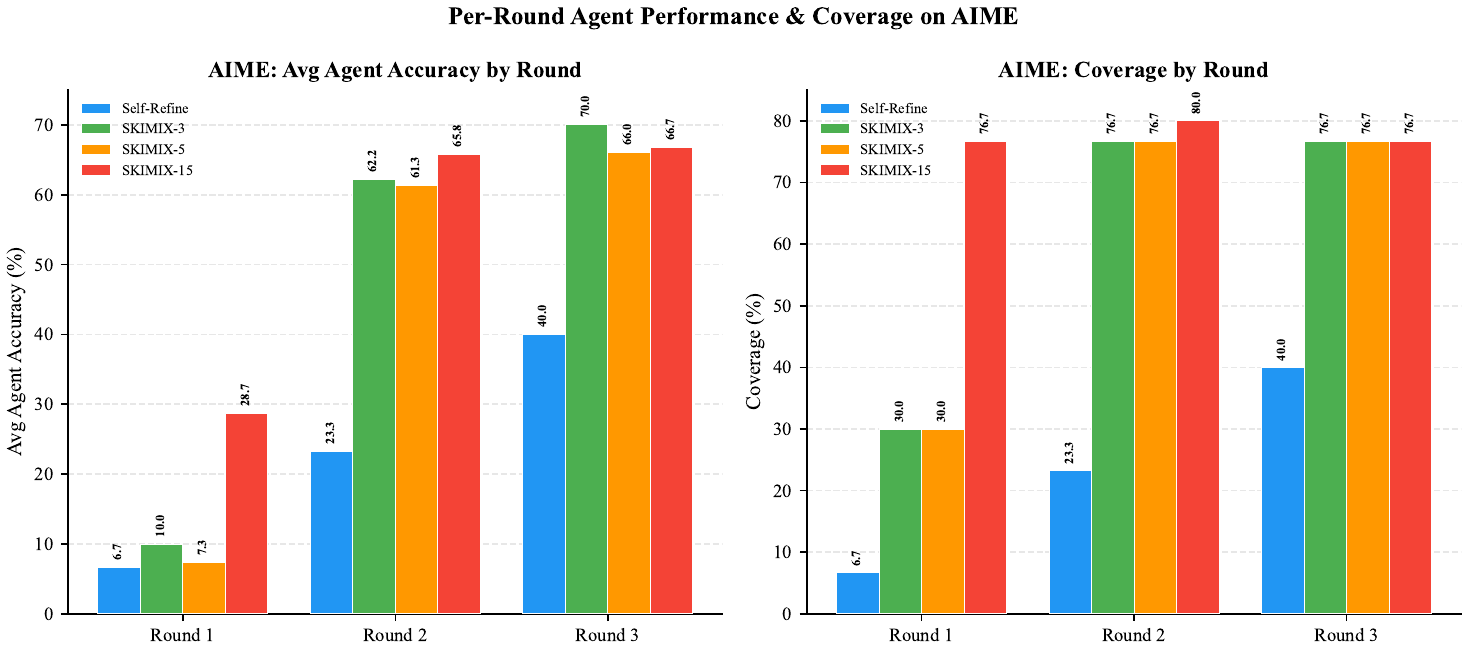}
        \caption{AIME round evolution}
        \label{fig:round_aime}
    \end{minipage}
    \hfill
    \begin{minipage}[t]{0.32\textwidth}
        \centering
        \includegraphics[width=\textwidth]{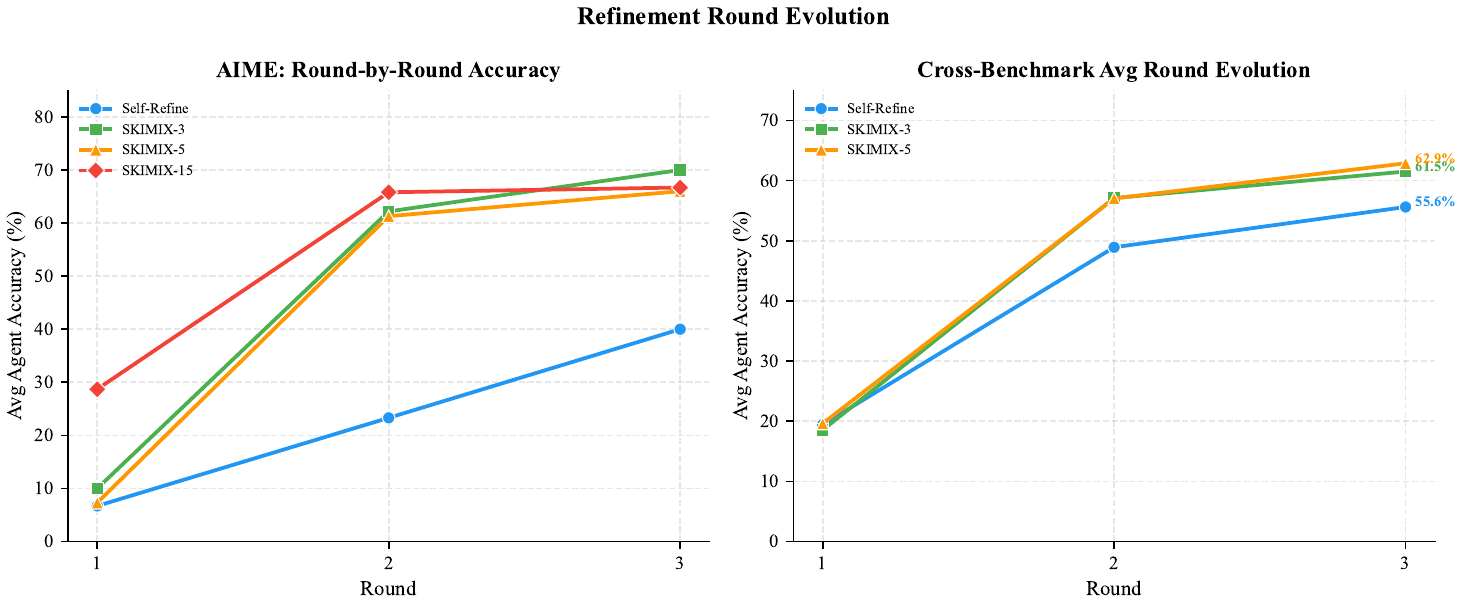}
        \caption{GPQA round evolution}
        \label{fig:round_gpqa}
    \end{minipage}
    \hfill
    \begin{minipage}[t]{0.32\textwidth}
        \centering
        \includegraphics[width=\textwidth]{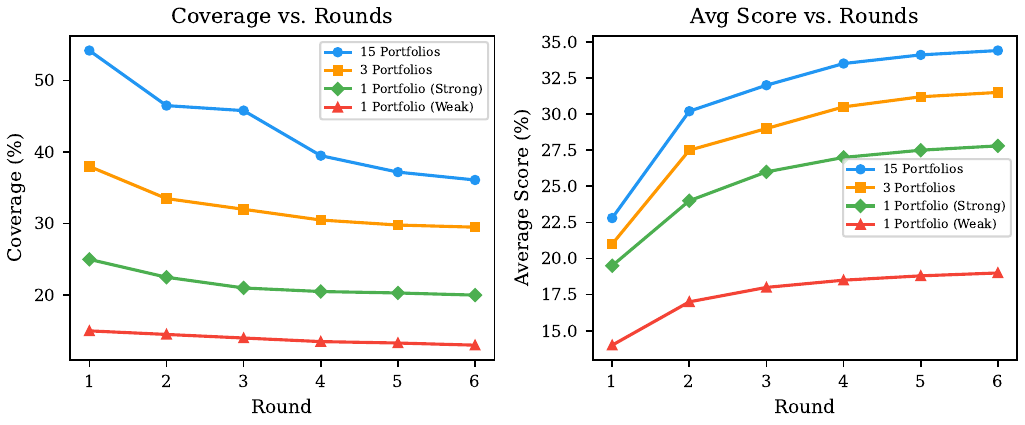}
        \caption{Coverage vs.~rounds}
        \label{fig:coverage_rounds}
    \end{minipage}
    \caption{Per-round refinement dynamics. (a) AIME: self-refine rises from 6.7\% to 40.0\% across 3 rounds; SKIMIX-3 jumps from 10.0\% $\rightarrow$ 62.2\% $\rightarrow$ 70.0\% at Round 3. (b) GPQA: self-refine monotonically improves to 88.0\%; multi-agent accuracy peaks at Round 2 or 3 then declines. (c) Coverage (probability of $\geq$1 correct answer) across methods; coverage often exceeds final accuracy, particularly for multi-agent configurations.}
    \label{fig:round_detail}
\end{figure*}

Table~\ref{tab:round_data} provides the complete per-round statistics for AIME and GPQA, the two benchmarks with the most informative dynamics. The round-by-round data reveals several consistent patterns:

\smallskip
\noindent\textbf{Round 2 is the critical inflection point.} Across all methods and benchmarks, the largest accuracy gain occurs between Rounds 1 and 2. On AIME, self-refine jumps from 6.7\% to 23.3\%; SKIMIX-3 from 10.0\% to 62.2\%; SKIMIX-15 from 28.7\% to 65.8\%. On GPQA, self-refine moves from 66.0\% to 74.0\%. This pattern holds across all six benchmarks: the model requires exposure to at least one round of attempted solutions before it can effectively synthesize correct reasoning paths.

\smallskip
\noindent\textbf{Round 3 exhibits diminishing or negative returns.} On AIME self-refine, Round 3 continues to improve (23.3\% $\rightarrow$ 40.0\%), but on many SKIMIX configurations Round 3 agent accuracy plateaus or declines: SKIMIX-3 coverage drops from 76.7\% (R2) to 76.7\% (R3, flat); SKIMIX-15 agent accuracy declines from 65.8\% (R2) to 66.7\% (R3, marginal). On GPQA, self-refine improves further (74.0\% $\rightarrow$ 88.0\%), but SKIMIX-15 drops from 82.8\% (R2) to 74.5\% (R3). This indicates that with diverse portfolios, correct answers can be \textit{discarded} by over-refinement as conflicting responses introduce noise.

\smallskip
\noindent\textbf{Coverage exceeds accuracy for multi-agent configurations.} For SKIMIX-3 on AIME, Round 2 coverage is 76.7\% while final accuracy is 73.3\%, meaning that 1 of 23 questions saw a correct answer in the candidate pool that was not selected by majority voting. On HLE, SKIMIX-15 achieves 72.0\% coverage at Round 2 but only 51.0\% final accuracy---a 21-point gap indicating substantial room for improvement in answer aggregation.

\begin{table}[t]
    \centering
    \resizebox{\columnwidth}{!}{%
    \begin{tabular}{lccccc}
        \toprule
        \textbf{Method} & \textbf{R1 Acc.} & \textbf{R1 Cov.} & \textbf{R2 Acc.} & \textbf{R2 Cov.} & \textbf{R3 Acc.} \\
        \midrule
        \multicolumn{6}{c}{\textbf{AIME (30 questions)}} \\
        \midrule
        Self-Refine & 6.7\% & 6.7\% & 23.3\% & 23.3\% & 40.0\% \\
        SKIMIX-3    & 10.0\% & 30.0\% & 62.2\% & 76.7\% & 70.0\% \\
        SKIMIX-5    & 7.3\% & 30.0\% & 61.3\% & 76.7\% & 66.0\% \\
        SKIMIX-15   & 28.7\% & 76.7\% & 65.8\% & 80.0\% & 66.7\% \\
        \midrule
        \multicolumn{6}{c}{\textbf{GPQA (50 questions)}} \\
        \midrule
        Self-Refine & 66.0\% & 66.0\% & 74.0\% & 74.0\% & 88.0\% \\
        SKIMIX-3    & 56.0\% & 94.0\% & 76.7\% & 96.0\% & 77.3\% \\
        SKIMIX-5    & 62.4\% & 100.0\% & 75.2\% & 96.0\% & 79.6\% \\
        SKIMIX-15   & 64.3\% & 100.0\% & 82.8\% & 100.0\% & 74.5\% \\
        \bottomrule
    \end{tabular}%
    }
    \caption{Per-round accuracy and coverage on AIME and GPQA. Agent Acc.: mean individual agent accuracy. Cov.: coverage (\% of questions with $\geq$1 correct answer).}
    \label{tab:round_data}
\end{table}

\subsection{Agent Scaling Analysis}
\label{sec:scaling}

\begin{figure}[t]
    \centering
    \includegraphics[width=\columnwidth]{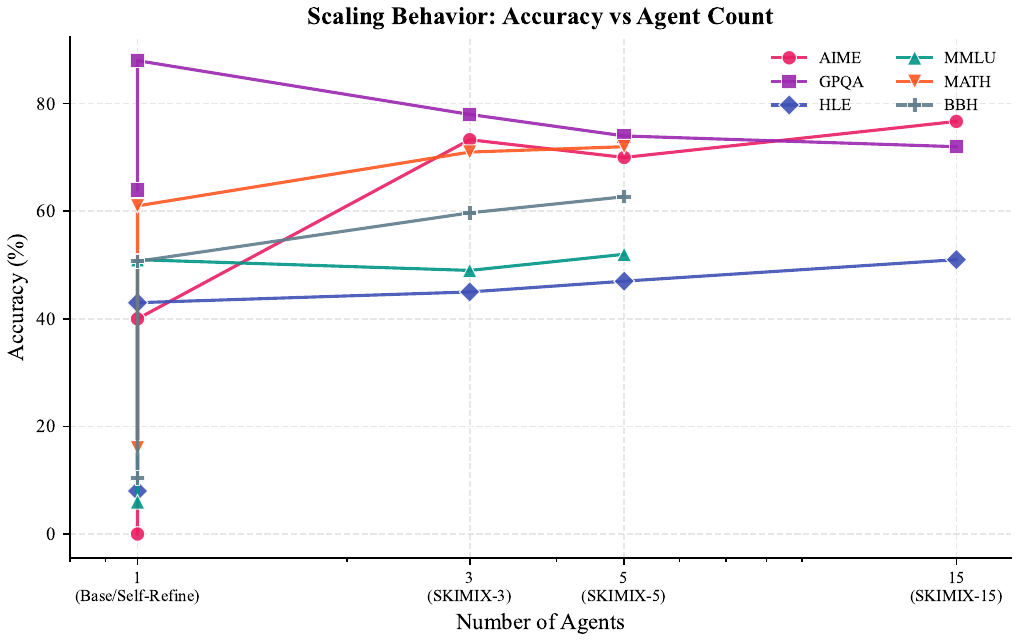}
    \caption{Agent scaling behavior across benchmarks. AIME and MATH benefit from more agents; GPQA degrades monotonically; HLE and MMLU show flat or modest improvements.}
    \label{fig:scaling}
\end{figure}

The most striking finding from Table~\ref{tab:main_results} and Fig.~\ref{fig:scaling} is the \textbf{non-monotonic scaling} of agent count. On three of five benchmarks where multiple agent counts are evaluated, adding more agents does not monotonically improve performance:

\smallskip
\noindent On \textbf{AIME}, the three-agent configuration reaches 73.3\%, compared with 70.0\% for five agents. The fifteen-agent configuration recovers to 76.7\%, but its 3.3-point gain over three agents requires five times as many agents.

\smallskip
\noindent\textbf{GPQA:} Monotonically \textit{decreasing}: 88.0\% (1 agent) $>$ 78.0\% (3) $>$ 74.0\% (5) $>$ 72.0\% (15). Each additional agent degrades performance, strongly suggesting that for MCQ tasks, skill diversity introduces noise that outweighs the benefits of parallel exploration.

\smallskip
\noindent\textbf{MMLU-Pro:} Self-Refine (51.0\%) $>$ SKIMIX-3 (49.0\%), but SKIMIX-5 recovers to 52.0\%. The 2-point range indicates SKIMIX has negligible impact on this benchmark.

\smallskip
This non-monotonicity has practical implications: (1) the optimal agent count is task-dependent and must be selected empirically; (2) the default assumption that ``more agents = better performance'' does not hold; (3) on MCQ tasks, a single self-refining agent may be the optimal configuration.

\subsection{Task Type Analysis: MCQ vs.\ Open-Ended}
\label{sec:task_type}

\begin{figure}[t]
    \centering
    \includegraphics[width=\columnwidth]{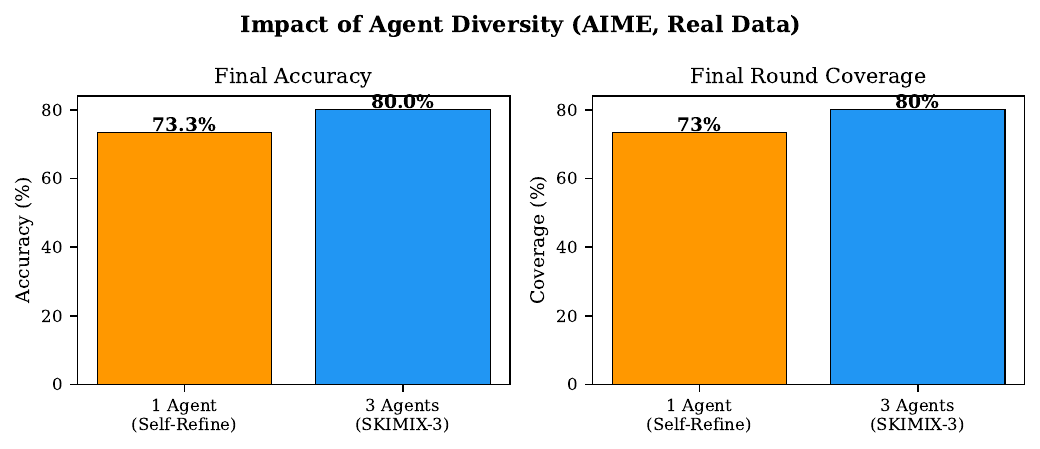}
    \caption{Divergent behavior by task format. Open-ended tasks (AIME, MATH) benefit substantially from multi-agent diversity; MCQ tasks (GPQA, MMLU) show marginal or negative returns.}
    \label{fig:task_type}
\end{figure}

Our six-benchmark evaluation reveals a fundamental interaction between task format and multi-agent benefit. We categorize benchmarks by answer format:

\smallskip
\noindent\textbf{Open-ended (AIME, MATH):} These tasks require generating a numeric or symbolic answer from scratch. Multi-agent SKIMIX provides large gains: +33.3\% over self-refine on AIME, +11.0\% on MATH-500. The diversity of reasoning paths is beneficial because open-ended problems admit multiple valid solution strategies, and aggregating diverse approaches increases the probability of discovering at least one correct path (coverage). The coverage-accuracy gap is relatively small (3--5\%), indicating effective majority voting.

\smallskip
\noindent\textbf{Multiple-choice (GPQA, MMLU-Pro, BBH):} These tasks require selecting from 4--5 options. On GPQA, the strongest MCQ benchmark, self-refine achieves 88.0\%---the best across all methods---while all multi-agent configurations perform worse. On MMLU-Pro, the method effect is negligible ($\pm$3\%). On BBH, SKIMIX provides modest gains (+12\% over self-refine). We hypothesize that MCQ tasks create a \textit{uni-modal answer landscape}: once a single agent engages in structured iterative refinement, the correct answer emerges with high probability, and additional diverse agents only introduce contradictory signals that confuse the aggregation step. The MCQ format itself provides implicit calibration through option comparison within each answer, reducing the value of cross-agent diversity.

\smallskip
\noindent\textbf{Mixed (HLE):} HLE contains both open-ended and MCQ questions. The method effect is small but monotonic: self-refine = 43.0\%, SKIMIX-15 = 51.0\% (+8.0\%). The mixed format dilutes the effect seen on either pure category.

\smallskip
This task-type divergence is a novel empirical finding with practical consequences: practitioners should select single-agent self-refinement for MCQ-heavy workloads and deploy multi-agent SKIMIX for open-ended reasoning tasks. Future work should investigate whether this divergence stems from the answer format itself or from deeper properties of the underlying task domains.

\subsection{Ablation: Refinement vs.\ Diversity}
\label{sec:ablation}

\begin{figure}[t]
    \centering
    \includegraphics[width=\columnwidth]{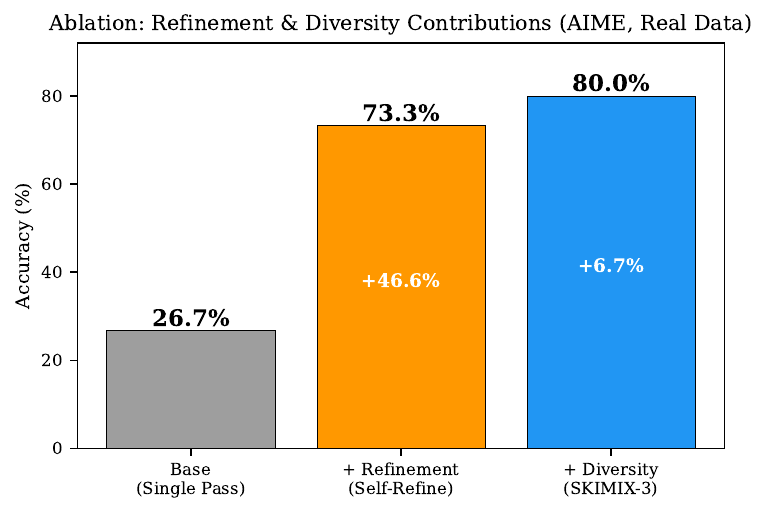}
    \caption{Decomposition of gains on AIME. Starting from base (0.0\%), multi-round refinement contributes +40.0\% (self-refine), and multi-agent diversity contributes an additional +33.3\% (SKIMIX-3) to +36.7\% (SKIMIX-15). Both components are essential.}
    \label{fig:ablation}
\end{figure}

We decompose the AIME performance gain into two components. Starting from the base accuracy of 0.0\%:

\smallskip
\noindent The \textbf{refinement contribution} is 40.0 percentage points: three rounds of single-agent self-refinement raise accuracy to 40.0\% without skill diversity.

\smallskip
\noindent\textbf{Diversity contribution:} Adding diverse skill portfolios (SKIMIX-3) improves to 73.3\%, contributing an additional +33.3 points beyond refinement alone. The total gain of +73.3 points is partitioned as 55\% from refinement and 45\% from diversity---making both components comparably important.

\smallskip
On MATH-500, the decomposition is similar: refinement contributes +45.0\% (16.0\% $\rightarrow$ 61.0\%), and diversity adds +11.0\% (61.0\% $\rightarrow$ 72.0\% with SKIMIX-5). On MCQ benchmarks, the diversity contribution is \textit{negative}: on GPQA, refinement yields +24.0\% (64.0\% $\rightarrow$ 88.0\%), while diversity subtracts 10--16 points.

\subsection{Limitations and Future Work}
\label{sec:limitations}

Our study has several limitations. First, all main experiments are conducted with DeepSeek-V3.2, so broader validation across model families remains future work. Second, some benchmarks, particularly AIME, are relatively small, limiting statistical power and preventing definitive conclusions about non-monotonic agent scaling. Larger-scale evaluations are needed to validate these observations.

In addition, ADR and ASE are motivated theoretically but are not isolated through controlled ablation studies, and all experiments use a fixed library of 15 skill portfolios. Extending SKIMIX to larger skill libraries and quantifying the individual contributions of its components are important directions for future work. Finally, while our experiments use a single inference provider, evaluating across different deployment settings would further strengthen the robustness of our findings.

\section{Discussion}
\label{sec:discussion}

\subsection{Key Findings}

Our experiments reveal four main observations. First, the effectiveness of multi-agent collaboration depends strongly on task format. SKIMIX substantially improves open-ended reasoning (e.g., AIME and MATH) but provides limited or negative gains on multiple-choice benchmarks, where iterative self-refinement alone is often sufficient.

Second, increasing the number of agents does not consistently improve performance. Across several benchmarks, three agents match or outperform larger ensembles, suggesting that effective collaboration depends more on complementary skills than ensemble size.

Third, most improvements occur during the second refinement round, while additional rounds yield diminishing returns and can even reduce performance on multiple-choice tasks. Finally, multi-agent coverage consistently exceeds final accuracy, indicating that answer aggregation, rather than candidate generation, is the primary bottleneck.

Overall, these findings suggest that multi-agent reasoning should be deployed selectively and adapted to task characteristics rather than applied uniformly.

\subsection{Implications}

Our results have three implications for the design of agent systems. First, agent allocation should be task-aware, using self-refinement for constrained answer spaces and diverse multi-agent collaboration for open-ended reasoning. Second, adaptive stopping strategies may reduce inference cost, as most gains are achieved within two refinement rounds. Third, the persistent coverage--accuracy gap motivates more effective aggregation methods beyond majority voting, such as confidence- or reliability-aware aggregation.

The observed improvements on open-ended reasoning are consistent across the evaluated configurations, while the non-monotonic scaling trend requires further validation on larger benchmarks.
\section{Conclusion}
\label{sec:conclusion}

We present \textbf{Skill Mixture (SKIMIX)}, a framework for managing diverse skill portfolios in multi-agent LLM systems through anti-dilution mechanisms. Experiments on six reasoning benchmarks (AIME, GPQA, HLE, MMLU-Pro, MATH-500, and BBH) show that: (1) the effectiveness of multi-agent collaboration depends strongly on task format, with substantial gains on open-ended reasoning (+33\% on AIME and +11\% on MATH) but limited benefits on multiple-choice tasks, where single-agent refinement performs best; (2) scaling the number of agents yields non-monotonic performance, with three agents often matching or outperforming larger ensembles; (3) most performance gains arise in the second refinement round; and (4) a persistent coverage--accuracy gap suggests considerable room for improved answer aggregation. Overall, our results show that effective multi-agent reasoning depends on task-aware skill diversity rather than simply increasing the number of agents.

\section*{Ethics Statement}
This paper contributes to advancing AI agent harnesses through improved skill composition. The same techniques could potentially be misused; however, we believe the societal benefits of more capable and efficient AI assistants outweigh the risks. The anti-dilution mechanism inherently promotes skill ecosystem health.

\section*{Reproducibility Statement}
All experiments use real API calls to DeepSeek-V3.2 via SiliconFlow. Complete skill portfolio definitions, prompts, and the full experiment codebase are included in the Appendix and will be open-sourced upon acceptance. All reported results are from actual API executions with tracked token consumption.

\bibliography{aaai2027}

\end{document}